\documentclass[11pt]{article}

\usepackage[]{EMNLP2022}

\usepackage{times}
\usepackage{latexsym}

\usepackage[T1]{fontenc}

\usepackage[utf8]{inputenc}

\usepackage{microtype}

\usepackage{inconsolata}

\usepackage{svg}

\title{End-to-End Speech to Intent Prediction to improve E-commerce Customer Support Voicebot in Hindi and English}

\author{Abhinav Goyal, Anupam Singh, Nikesh Garera \\
  Flipkart \\
  \texttt{\{abhinav.goyal,anupam.s,nikesh.garera\}@flipkart.com}}

\begin{document}
\maketitle
\begin{abstract}

Automation of on-call customer support relies heavily on accurate and efficient speech-to-intent (S2I) systems. Building such systems using multi-component pipelines can pose various challenges because they require large annotated datasets, have higher latency, and have complex deployment. These pipelines are also prone to compounding errors. To overcome these challenges, we discuss an end-to-end (E2E) S2I model for customer support voicebot task in a bilingual setting. We show how we can solve E2E intent classification by leveraging a pre-trained automatic speech recognition (ASR) model with slight modification and fine-tuning on small annotated datasets. Experimental results show that our best E2E model outperforms a conventional pipeline by a relative $\sim$27\% on the F1 score.

\end{abstract}

\section{Introduction}

Spoken Language Understanding (SLU) systems that extract the intent from a spoken utterance are integral in various voicebot applications such as automated on-call customer support, voice assistants, home or vehicle automation systems, etc. The extracted intent triggers a standard operating procedure (SOP) as defined by the respective application, e.g. an e-commerce customer query ``I want to return my phone'' maps to ``Return'' intent which triggers the SOP to help the user with returns. It helps us reduce the reliance on human agents and provide faster resolutions. More elaborate examples are shown in Table~\ref{tab:voicebot_examples}.

Conventionally, such systems consist of two components - an Automatic Speech Recognition (ASR) system followed by a Natural Language Understanding (NLU) unit. ASR converts audio to text, and NLU performs intent classification. Further, each component can have multiple sub-models. Typically, both these components are developed and optimized independently. ASR optimizes word error rate (WER) with equal weightage to individual words. This might not be optimal for an S2I system since all words are not equally relevant for intent classification. Also, due to the broad diversity in speech, training reliable ASR models can be very data intensive and strenuous. An error-prone ASR results in noisy inputs to NLU models, typically trained on clean text. This causes error accumulation which reduces the pipeline's performance. Data-intensive training of multiple models, high complexity \& maintenance, and higher overall latency make this pipeline approach sub-optimal.

The end-to-end S2I model is an intuitive alternative to overcome these limitations. It eliminates the problem of error accumulation, is simple and faster, and reduces the efforts required for independent models. Modelling the problem as audio-to-intent classification simplifies the task since the number of intents is usually much less than the vocabulary size used in ASR and NLU. It helps us reduce the requirement of manually annotated training data.

In this work, we adapt an E2E ASR model to build an E2E S2I model for Flipkart's on-call customer support. An overview of our contributions is as follows:

\begin{itemize}
    \item An efficient extension of end-to-end BiLSTM and CTC based ASR models for S2I task on noisy datasets;
    \item A demonstration of how the idea can outperform conventional pipeline in customer support voicebot in real-world settings;
    \item An investigation on how ASR pre-training, offline active learning and pseudo labelling reduce data labeling requirements for S2I.
\end{itemize}

Next, we discuss some related work in Section~\ref{literature}. Section~\ref{baseline_model} \& \ref{e2e_S2I} describe the baseline S2I pipeline and our E2E approach respectively. We talk about datasets, preprocessing and experimental setup in Section~\ref{experiments}. Finally, we conclude with a discussion on results and limitations in Section~\ref{results}.

\begin{figure}[htbp]
  \centering
  \includesvg[height=0.7\columnwidth]{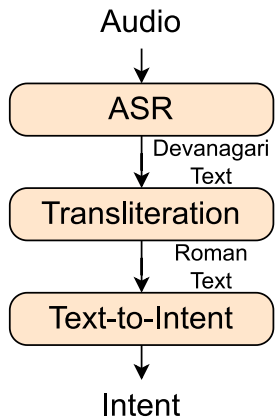}
  \caption{Text-based baseline system}
  \label{fig_baseline_model}
\end{figure}

\section{Related Work}
\label{literature}

There have been several attempts to mitigate ASR error propagation in text-based pipelines. One straightforward idea is to correct the ASR output, using error correction models \cite{asr_correction_nlu, asr_correction} or by ranking n-best hypotheses \cite{nbest3, nbest4, nbest5}. Other approach is to leverage extra information from ASR - output lattice \cite{latticernn,latticetf, lattice}, n-best hypotheses \cite{rerank_asr_nlu, nbest1, nbest2} or word confusion networks/embeddings \cite{word_confusion, confusion2vec}. Though these approaches make NLU robust to some ASR errors, they still use a strict multi-component pipeline.

There have been an increasing number of attempts toward building end-to-end SLU models. \citet{e2e1, e2e2, e2e3} investigate end-to-end SLU models which do not use ASR at all whereas \citet{e2e_joint} optimizes ASR and NLU in a joint setup. Such end-to-end models can require a large amount of paired speech and intent data which may not always be available. \citet{upt1, upt2} explore unsupervised pre-training which helps in low-resource settings but is usually very compute intensive. An alternative approach is to initialize SLU models using weights trained for ASR \cite{asr_slu2, asr_slu3, asr_slu1}. Since ASR datasets are more easily available, this approach presents a much easier method of pre-training than unsupervised methods.

Inspired by ASR pre-training, we explore how to augment a pre-trained ASR model for end-to-end S2I task for Flipkart customer support voicebot in Hindi and English languages.

\section{Text-based Pipeline}
\label{baseline_model}

Our baseline consists of 3 components - ASR, transliteration and text-to-intent as shown in Fig.~\ref{fig_baseline_model}. We use a bilingual ASR system which predicts text in Devanagari script for both Hindi and English. The transliteration model converts this text into Roman script. Finally, the text-to-intent model extracts intent from Roman text.

\begin{figure}[htbp]
  \centering
  \includesvg[width=0.92\columnwidth]{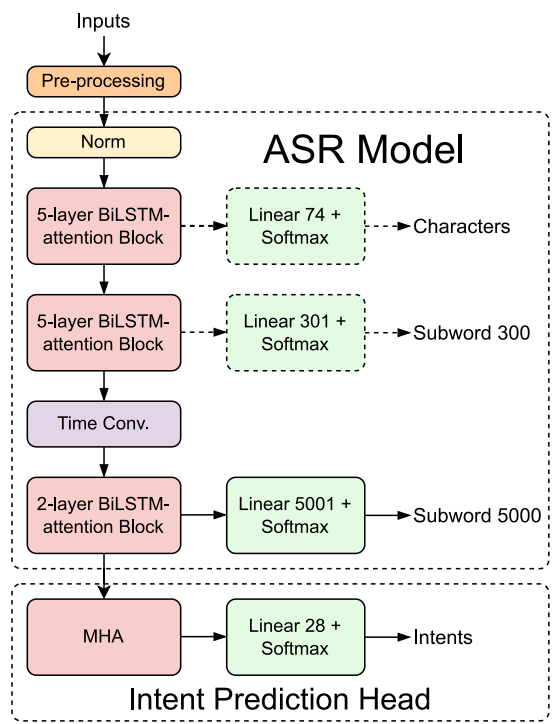}
  \caption{E2E Speech-to-Intent. Features from the last ASR block are used as inputs for intent classification.}
  \label{S2I_model}
\end{figure}

\subsection*{Automatic Speech Recognition}

\begin{figure}[htbp]
  \centering
  \includesvg[width=0.8\columnwidth]{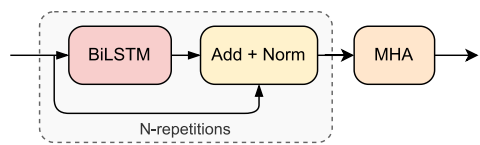}
  \caption{N-layer BiLSTM-attention block}
  \label{bilstm_block}
\end{figure}

Inspired by ~\citet{hctc_paper}, we use a 3-level HCTC architecture based on LSTM and attention \cite{transformer} as shown in Fig.~\ref{S2I_model}. Going in a fine-to-course fashion, the model predicts characters (73 tokens), short subwords (300 tokens) and long subwords (5000 tokens) at the respective levels. We use unigram models from Sentencepiece \cite{sentencepiece} for text segmentation. Each level consists of an N-layer LSTM-attention block (Fig.~\ref{bilstm_block}), N being 5-5-2, followed by a linear softmax layer.

\noindent For inference, the output of the last block, with 5000 subword units, is used for decoding the text using prefix beam search. The top 100 candidates are then re-ranked using a 3-gram KenLM \cite{kenlm} to select the best one.

\subsection*{Transliteration}
\label{transliteration}

A manually curated mapping and a fallback transformer encoder-decoder model \cite{transformer}, with a single layer each in encoder and decoder, is used for transliteration. The transformer uses a sum of character and position embeddings as inputs. Together, this combination has a WER of <1\% on unique utterances from a blind test set.

\subsection*{Text-to-Intent}

For text-to-intent classification, we have 28 categories (26 intents + others + blank) related to different customer queries, e.g. ``Delivery status'', ``Product return'' etc. We use the ``Blank'' intent when the output text is blank. For the baseline, we try different models, out of which XGBoost \citep{xgboost} with TF-IDF features gives the best results. We observe that neural network-based models - BiLSTM and BERT \citep{bert} overfit on our dataset. BERT when pre-trained on a large corpus performs at par with XGBoost.

\section{E2E Speech-to-Intent}
\label{e2e_S2I}

For the S2I task, we augment the pre-trained ASR model (same as used in the baseline) with intent prediction head as shown in Fig.~\ref{S2I_model}. We summarize the hidden features from the last block of the ASR model using a dot-product based multi-headed self-attention (MHA) layer. We use the output sequence of the last block as key-value vectors and the final cell state of the last BiLSTM layer as the query vector. A linear layer then predicts probability distribution over the intent classes. Since there's no text output from the model, the "Blank" intent is also predicted the by E2E S2I model. We train the intent prediction head (and fine-tune the BiLSTM blocks) using cross entropy loss.

\begin{table*}
\centering
\begin{tabular}{lllr}
\hline
\textbf{Task} & \textbf{Model} & \textbf{Source} & \textbf{Rough Size of Dataset}\\
\hline
ASR & BiLSTM & Voicebot & 893 hrs \\
& & Voice Search & 9.9k hrs \\
& & Generic & 6.4k hrs \\
& SPM, KenLM & Voicebot & 920k sentences \\
& & Others & 10M sentences \\
\hline
Transliteration & Transformer & Generic & 96k word pairs \\
\hline
Text-to-Intent & all & Voicebot & 55k text-intent pairs \\
& & Chatbot & 35k text-intent pairs \\
\hline
Speech-to-Intent & V1 & Voicebot & 10k audio-intent pairs \\
& V2 & Voicebot & 10k+25k audio-intent pairs \\
\hline
\end{tabular}
\caption{Breakup of various datasets used for training.}
\end{table*}

\section{Experiments}
\label{experiments}

\subsection{Datasets}

\subsubsection*{Automatic Speech Recognition}

A collection of datasets is used to train the ASR model - Flipkart customer support voicebot queries, voice search queries and general domain speech data. We transcribe all the utterances using an existing ASR system and manually correct the errors. The ASR system used to generate reference text is incrementally improved as more data is available. There's no control over the recording environment, and the correction of ASR transcripts instead of transcription from scratch leaves some errors introduced by the ASR model. This causes the dataset to have a lot of acoustic and textual noise. The datasets collectively amount to $\sim$11 M audio-text pairs which correspond to roughly 17k hours of audio. It has a mix of Hindi and English (possibly code-mixed) languages.

We train KenLM and Sentencepiece on a large corpus collected from various sources such as Flipkart's customer support chatbot and voicebot, voice search queries and product catalogue. The $\sim$920k Voicebot utterances (in-domain data) are upsampled during training.

\subsubsection*{Transliteration}

The transformer model is trained on $\sim$96k unique words which are manually transliterated. This dataset consists of high frequency words in Hindi and English in equal proportions. We add manual transliterations of words frequent for our use case in the look-up dictionary.

\subsubsection*{Text-to-Intent}

For training text-to-intent, $\sim$90k manually labeled unique text-intent pairs are used. This mainly consists of customer support voicebot and chatbot queries. For deep neural-network based models - BiLSTM and BERT, a large pre-training corpus from e-commerce domain is also used.

\subsubsection*{Speech-to-Intent}

For fine-tuning the model for S2I, we use a set of 10k randomly sampled voicebot queries (we call it V1) and manually label the intents. We also use additional 25k audios for offline active learning. We name this complete 10k+25k set as V2. 

Since the legacy system uses independent models, training data, to be annotated, was sampled independently and randomly for each model. The training datasets for ASR and Text-to-Intent don't have a large intersection. Therefore, we don't have a large dataset for training E2E S2I models and instead use smaller, independently labelled datasets.

\subsection{Pre-processing and Experimental Setup}

We use standard log-mel-spectrogram features with a window of 20ms, a stride of 10ms, and FFT size of 512. The number of filterbanks used is 80. We use masking \cite{specaugment} for data augmentation. We also stack 5 consecutive frames with a stride of 3 frames giving an input feature vector of 400 size with a receptive field of 60ms and stride of 30ms for each time step.

We use a cyclical learning rate (LR) \cite{cyclic_lr} to train the ASR model for 8 epochs with a batch size of $\sim$42 minutes. For S2I, we use constant LR, batch size of $\sim$26 minutes and fine-tune it in 2 steps - on V1 dataset for 10 epochs + V2 dataset for 6 epochs. Training takes $\sim$2.5 days for ASR and $\sim$24 minutes for S2I on 1 A100 GPU.

\section{Results}
\label{results}

\begin{table*}
  \centering
  \begin{tabular}{lccccccc}
  \hline
  & \textbf{\#params}&& \multicolumn{2}{c}{\textbf{All Intents}} & \multicolumn{2}{c}{\textbf{Except blank/other}}\\
  \textbf{Model} & \textbf{(in M)} & \textbf{WER} & \textbf{Accuracy} & \textbf{F1 score} & \textbf{Accuracy} & \textbf{F1 score}\\
  \hline
  Baseline & 48.60 & 8.34 & 83.62 & 82.78 & 82.76 & 85.36 \\
  Baseline with GT text & - & 0 & 86.22 & 85.84 & 82.01 & 84.76 \\
  \hline
  S2I linear (V1) & 43.85 & - & 86.22 & 85.85 & 84.35 & 87.08 \\
  S2I MHA (V1) & 45.97 & - & 86.28 & 85.85 & 85.88 & 87.81 \\
  S2I MHA from scratch (V1) & 45.97 & - & 70.06 & 69.31 & 60.69 & 61.17 \\
  S2I MHA (V2) & 45.97 & - & 87.18 & 87.13 & \textbf{87.60} & \textbf{89.94} \\
  \textbf{S2I MHA (V2+pseudo lab.)} & 45.97 & - & \textbf{87.49} & \textbf{87.37} & 87.00 & 89.57 \\
  \hline
  \end{tabular}
  \caption{Results on Intent Prediction. ``Baseline'' is the text based pipeline where text is given by the ASR system. ``Baseline with GT text'' is where we substitute ASR with true transcriptions. All numbers are in \%.}
  \label{tab:S2I_results}
\end{table*}

We compare the baseline and E2E model on 14606 voicebot queries manually transcribed and annotated for text and target intent resp. We report accuracy and F1 score for intent classification and word error rate (WER) for ASR in Table~\ref{tab:S2I_results}. The ASR system used for baseline has a WER of 8.34\%. As mentioned earlier in Section~\ref{transliteration}, transliteration module has a WER of <1\%. Together, the WER of the ASR + transliteration system becomes <9.2\%. The text-to-intent model has an F1 score of 85.84\%. We compute this using manual transcriptions as inputs to the text-to-intent model.

The S2I model, fine-tuned on just 10k manually annotated audio-intent pairs (V1), outperforms the baseline by an absolute 3.07\% on the F1 score. Using this, we predict intent on an unlabeled set and get a random sample of 25k audios where the model has low confidence (prediction probability as given by softmax on the last layer). We correct this set manually and re-train the model using all 35k samples (V2), improving the F1 score by 1.28\%. We then re-train the model on the complete set of voicebot queries ($\sim$920k audios from ASR dataset) using pseudo labels, further improving the score marginally. Our final E2E model outperforms the baseline by an absolute 4.59\% on the F1 score.

The E2E model has a median latency of $\sim$41ms, which is 1/3rd of the baseline latency ($\sim$123ms). Since we can deploy the complete model on a GPU, it can handle inference at a much larger scale than the baseline - more than 1000 queries per second using a single A100 GPU. Whereas the decoder in the ASR system used for the baseline, which is the bottleneck, can only handle about 90 queries per second. Thus, the E2E model outperforms the baseline on accuracy, latency, and scalability.

\subsection{Analysis and Discussion}

We also evaluate a simple time average of sequence output from the last ASR block in place of MHA. It gives almost the same results as MHA showing that the ASR model can adapt to intent classification task without extra modelling efforts. We observe that training the S2I model from scratch performs very suboptimal, which shows the importance of initializing the network using the ASR task when paired audio-intent data is scarce.

\begin{table*}
  \centering
  \begin{tabular}{|l|p{3.3cm}|p{3.4cm}|p{2.2cm}|p{2.2cm}|p{2.2cm}|}
  \hline
  \textbf{\#} & \textbf{Utterance} & \textbf{ASR output} & \textbf{True intent} & \textbf{Baseline} & \textbf{Ours}\\
  \hline
  1 & \textit{Wapas} karne ka hai & \textit{Wapis} karne ka hai & Return & Others & \textbf{Return}\\\hline
  2 & Das \textit{din} ke ander mujhe delivery chahiye & Das \textit{June} ke ander mujhe delivery chahiye & Specific delivery time & Delivery info & \textbf{Specific delivery time}\\\hline
  3 & Meri \textit{watch} khrab hai & Meri \textit{bahut} khrab hai & Return & Others & \textbf{Return}\\\hline
  4 & Ji zaroor kariye & Ji zaroor kariye & Yes & Others & \textbf{Yes}\\\hline
  5 & Delivery \textit{ki} timing & Delivery \textit{ke} timing & Delivery time & \textbf{Delivery time} & Delivery info\\\hline
  6 & Kuchh nhi haan haan & Kuchh nhi haan haan & End & \textbf{End} & Yes\\\hline
  \end{tabular}
  \caption{Speech-to-Intent examples. In 1-4, our model does better and in 5 \& 6, baseline does better.}
  \label{tab:result_examples}
\end{table*}

The text-to-intent model has a higher F1 score than the complete pipeline (85.84\% vs 82.78\%), suggesting that errors by the ASR model are the reason for the baseline's suboptimal performance. Our S2I model is not only able to mitigate this but also gives more improvement as it is 1.53\% better than standalone text-to-intent with manual transcriptions. Table~\ref{tab:result_examples} shows some examples to compare our S2I model with the baseline. In examples 1-3, ASR makes a mistake due to wrong pronunciation in 1 and high background noise in 2 \& 3. These errors cause the text-to-intent model to give wrong predictions demonstrating how error propagation affects the pipeline. In example 4, the intent model makes an error even with the correct transcription. In examples 5 and 6, the baseline outputs the right intent but the E2E model makes mistakes. In both cases, the E2E model confuses the intent with another very close category.

\subsection{Conclusions}

In this work, we show that pre-trained CTC-based end-to-end ASR models can be adapted for end-to-end Speech-to-Intent classification with slight augmentation and relatively much less annotated data. Our S2I model outperforms the text-based pipeline by an absolute 3.07\% on the F1 score while keeping the model size small and requiring only 10k annotated audio-intent pairs to train. It also simplifies the pipeline by eliminating the requirement of a dedicated ASR decoder, Text-to-Intent model, and language models. With just 25k additional labelled training pairs, our final model is $\sim$27\% better than the baseline on the F1 score (absolute improvement of 4.59\%). Thus, we show that the E2E S2I model, adapted from ASR, outperforms the conventional pipeline on accuracy, latency, and scalability while requiring much less labelled training data, compute resources, and modelling efforts.

\section*{Limitations}

The baseline text-to-intent model was trained on a different dataset from what is used for fine-tuning the E2E models. But, using a considerably smaller dataset than the baseline system puts the E2E models at a disadvantage. Since the E2E models still outperform the baseline, the absolute difference in utterances should not affect the credibility of the results. We also need a transliteration module in the text-based pipeline because of a mismatch in scripts used by the ASR and Text-to-Intent models. One might argue that the transliteration module weakens the baseline. But being a highly accurate system (with <1\% WER), it should not affect the baseline much. We have also verified this with a manual analysis of errors. Also, the ASR pre-training uses a large dataset which might not be readily available for low-resource languages.

As the future steps, it would be interesting to see how such models would work with more than two languages, where some of the languages are low-resource from ASR perspective. In such a setting, the ASR models usually employ a larger vocabulary to accommodate all languages, but since intent categories are the same, adapted S2I models can reduce vocabulary size by a greater extent giving them an even bigger advantage.

\begin{table*}[htbp]
  \centering
  \begin{tabular}{|l|p{3.5cm}|p{2.5cm}|p{8cm}|}
  \hline
  \textbf{\#} & \textbf{Customer Utterance} & \textbf{Intent/SOP} & \textbf{Voicebot Response}\\
  \hline
  \hline
  & & Welcome & Namaste, aapka svagat hai. Main aapki automated support assistant.
  (\textit{English: Hello and welcome. I am your automated support assistant}) \\\hline
  \hline
  1 & & Order identification & Kya aap apne “curtain” ke bare mein baat karna chahte hain, jo “shipped” hai?
  (\textit{English: Do you want to ask about your “curtain”, which is “shipped”}) \\\hline
  & Haan \textit{(English: Yes)} & Yes/Confirmation & Aapka order ship ho chuka hai, aur aapko aaj mil jayega. Aapko kya madad chahiye? \textit{(English: Your order has been shipped and you will get it by today. What do you need help with?)} \\\hline
  & Mera order aaj kitne time tak ghar pe aa jayega \textit{(English: By what time today will my order reach?)} & Delivery Info & Hamare agent aaj shaam 6 bje tak deliver karne ki koshish karenge. Yadi nahi ho paya to agle karyadin deliver karenge. \textit{(English: Our agents will try to deliver it by 6 pm today. If not possible, they will deliver it the next working day.)} \\
  \hline
  2 & & Order identification & Kya aap “usha air cooler” ke bare poochhna chahte hain, jo “returned” hai? \textit{(English: Are you calling about “Usha Air Cooler”, which is “returned”)} \\\hline
  & Yes & Yes/confirmation & What do you need help with? \\\hline
  & When will I get my replaced one cooler? & Return status & Your replacement will be delivered within 2 days after pickup of returned product. \\
  \hline
  3 & & Order identification & Aapne haal hi mein ek se zyada order place kiye hain. Aaj apne kis order ke bare mein call kiya hai? \textit{(English: Recently you placed more than 1 order. Which order are you calling about?)} \\\hline
  & Integriti jeans & & Apka “integriti jeans” ka order delivered hai. Aapko is order ke bare mein kya madad chahiye? \textit{(English: Your order of “integriti jeans” has been delivered. What help do you need with this order?)} \\\hline
  & Customer executive se baat karwao. \textit{(English: Let me talk to a customer executive.} & Agent request & Filhal hamare customer support executive vyast hain, aapse subah 9:51 sampark karenge. Kya main call ki vyavastha kar du? \textit{(English: Currently our executives are busy. They could call you at 9:51 am in the morning. Should I arrange for a call back?)} \\\hline
  & Haan thik hai. \textit{(English: yes it’s fine.)} & Yes/Confirmation & Thik hai, maine call ki vyavastha kar di hai. \textit{(English: Ok, I have arranged for a call back.)} \\
  \hline
  4 & & Order identification & Kya aapne TV ke bare mein call kiya hai? \textit{(English: Have you called about TV?)}\\\hline
  & Haan ji \textit{(English: Yes)} & Yes/Confirmation & Aapka order ship ho chuka hai aur 19 November tak deliver ho jaega. Aapko iss bare mein kya madad chahiye? \textit{(English: Your order has been shipped and will be delivered by 19th November. What help do you need regarding this?)}\\\hline
  & Please deliver it today itself & Faster delivery request & Sorry, while faster delivery is not available, please be assured we will deliver your order on time.\\\hline
  \hline
  & & Conversation end & *** mein call karne ke liye dhanyavaad, apka din shubh rahe. \textit{(English: Thanks you for calling ***, have a good day.)} \\
  \hline
  \end{tabular}
  \caption{Illustrations of how an S2I system can help provide faster and automated resolutions in e-commerce.}
  \label{tab:voicebot_examples}
\end{table*}

\bibliography{anthology,custom}
\bibliographystyle{acl_natbib}

\end{document}